%% file: iclr2026_conference.tex
\PassOptionsToPackage{table}{xcolor}
\documentclass{article} 
\usepackage{iclr2026_conference,times}

\input{math_commands.tex}

\usepackage{hyperref}
\usepackage{url}
\usepackage{graphicx}
\usepackage{xcolor}
\usepackage{makecell}
\usepackage{multirow}
\usepackage{wrapfig}

\definecolor{secondblue}{RGB}{45, 118, 185}
\newcommand{\best}[1]{\textcolor{red}{\textbf{#1}}}
\newcommand{\second}[1]{\textcolor{secondblue}{\textbf{#1}}}

\title{SegDem: Segmentation helps Demosaicing}

\author{
Ping Chen$^{1}$,
Xiangming Wang$^{1}$,
Yongyong Chen$^{1}$,
Jiezhang Cao$^{2}$,
\AND
Kai Zhang$^{3}$,
Jingyong Su$^{1}$,
Jie Liu$^{1}$,
Haijin Zeng$^{1}$\thanks{Corresponding author.} \\[0.5em]
{\normalfont $^{1}$Harbin Institute of Technology, Shenzhen} \\
{\normalfont $^{2}$Shanghai Jiao Tong University} \\
{\normalfont $^{3}$Nanjing University (Suzhou)} \\
{\normalfont \texttt{pingchen2004@gmail.com}, \texttt{haijin.zeng2018@gmail.com}}\\
{\normalfont Code will be available at \url{https://github.com/cipi666/SegDem}.}
}

\iclrfinalcopy
\begin{document}

\maketitle
\lhead{Preprint}


\begin{figure*}[htbp]
  \centering
  \includegraphics[width=\textwidth]{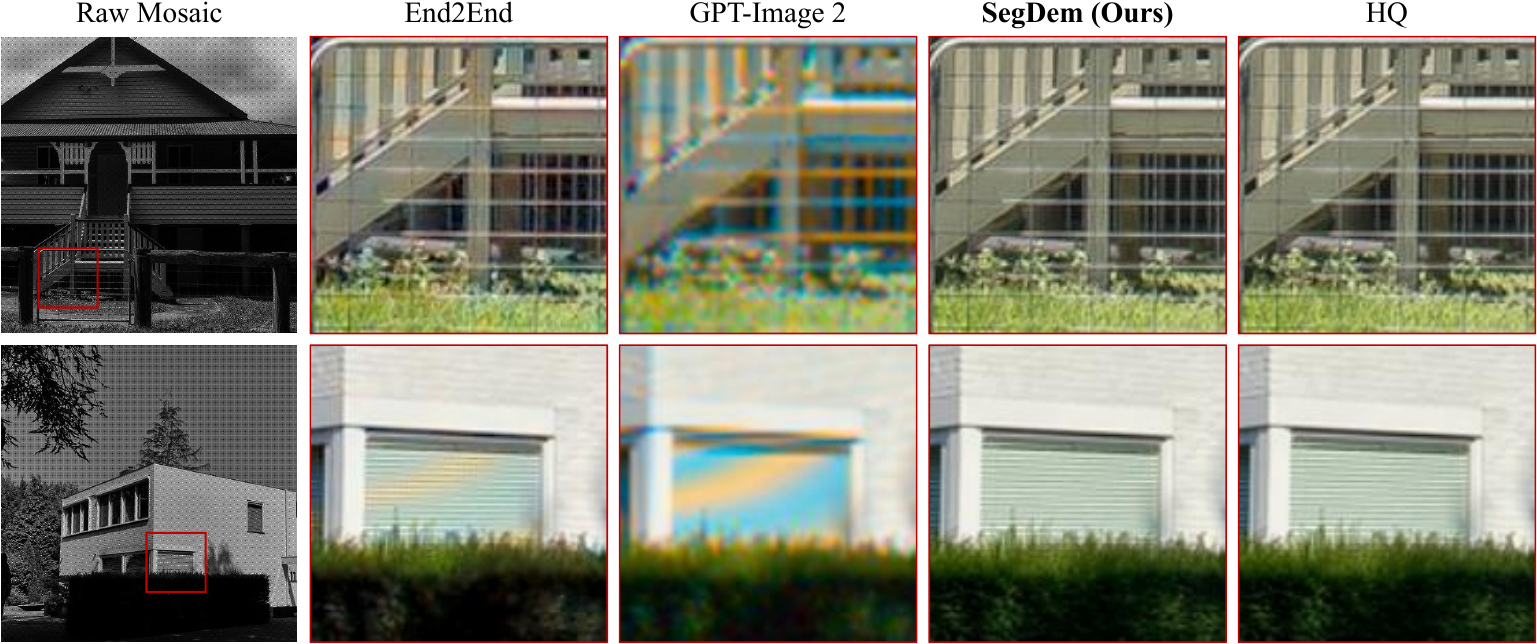}
  \caption{\textbf{Motivation of SegDem.} Conventional regression is prone to false colors and moir\'e artifacts in ambiguous regions, while generative models may hallucinate details unsupported by RAW measurements. SegDem instead transfers structural decoder representations from segmentation to demosaicing, improving boundary-aware reconstruction.}
  \label{fig:introduction}
\end{figure*}

\begin{abstract}
Image demosaicing reconstructs a full-color image from incomplete color measurements produced by a sensor covered with a color filter array (CFA). Most existing methods formulate demosaicing as pixel-level reconstruction and mainly rely on local textures, cross-channel correlations, and low-level image statistics. Our core insight is that reconstruction and visual understanding can be viewed as complementary views of shared scene structure: both are grounded in the same underlying physical world, and therefore the structural and physical information inferred from an image should remain consistent across the two tasks. We instantiate this idea with instance segmentation and propose \emph{SegDem}, a cross-task decoder representation transfer framework for demosaicing. SegDem first learns region- and boundary-aware representations through instance-aware structural pretraining and then transfers the decoder to RAW-conditioned reconstruction. Segmentation- and demosaicing-conditioned features are further anchored to a shared frozen DINOv2 representation space to preserve structural organization across tasks. We instantiate SegDem with convolutional, Transformer-based, and state-space backbones for unified Single- and Quad-Bayer demosaicing. Extensive experiments on synthetic, external, and challenging datasets demonstrate consistent improvements across different architectures and CFA layouts.
\end{abstract}

\section{Introduction}
\label{sec:introduction}

Image demosaicing reconstructs a full-color linear RGB image from spatially incomplete measurements acquired through a color filter array (CFA)~\citep{bayer1976color,DBLP:journals/ijcv/ZengFCHZLAP24}. Learning-based methods typically perform end-to-end regression from CFA observations to RGB images~\citep{malvar2004high,gharbi2016deep,kokkinos2018deep,DBLP:conf/iccp/TedlaPZB25,DBLP:conf/iccv/TedlaLYAB25,DBLP:conf/cvpr/ZhouZLSTCLS25}. Although effective, pixel-level reconstruction remains ambiguous around object boundaries, thin structures, and repetitive patterns, where multiple local color continuations may be consistent with the same CFA observation. Consequently, existing methods can still produce false colors, zipper artifacts, and moir\'e patterns, particularly when handling CFA layouts with different sampling geometries, such as Single-Bayer and Quad-Bayer patterns~\citep{malvar2004high,kokkinos2018deep,lee2023efficient,DBLP:conf/iccp/TedlaPZB25}.

Recent large-scale generative models have shown that high-level perceptual priors can improve low-level restoration tasks~\citep{wang2023stablesr,lin2023diffbir,wu2024seesr,yu2024supir,DBLP:journals/corr/abs-2603-25502,DBLP:journals/corr/abs-2604-03061,DBLP:conf/cvpr/ZengWCSL25}. However, directly applying generative synthesis to demosaicing is problematic because the reconstructed image should remain anchored to RAW measurements and CFA sampling. Generative restoration may hallucinate textures or shift colors unsupported by the sensor observation~\citep{blau2018perception,yin2026howfar,DBLP:journals/corr/abs-2604-03061}, as illustrated in Figure~\ref{fig:introduction}. This motivates us to reconsider the relation between visual understanding and reconstruction. Rather than treating them as isolated objectives, both rely on recovering and representing coherent scene structure: understanding organizes visual content into meaningful regions and boundaries, while faithful reconstruction should recover the structures that make such understanding possible.

From this perspective, high-level understanding can provide structural cues for resolving ambiguous low-level measurements. We use instance segmentation as a concrete instantiation of visual understanding, since it explicitly captures object regions and boundaries~\citep{he2017mask}. Rather than introducing segmentation prediction into demosaicing, we transfer segmentation-pretrained decoder representations to reconstruction, allowing structural knowledge to guide color recovery while preserving RAW measurement fidelity.

Based on this insight, we propose \emph{SegDem}, a cross-task decoder representation transfer framework for unified Single- and Quad-Bayer demosaicing. Stage~1 learns region- and boundary-aware decoder representations through instance-aware supervision, and Stage~2 transfers them to RAW-conditioned reconstruction. To preserve structural consistency across tasks, segmentation- and demosaicing-conditioned decoder features are aligned to a shared frozen DINOv2 representation space~\citep{oquab2023dinov2}. Generally, we instantiate SegDem with convolutional~\citep{You_2025_ICCV}, Transformer-based~\citep{zamir2022restormer}, and state-space!\citep{guo2024mambair} backbones.

Our contributions are summarized as follows:
\begin{itemize}
    \item We introduce a cross-task perspective on demosaicing that connects visual understanding and reconstruction through shared structural representations.
    
    \item We propose SegDem, which transfers region- and boundary-aware decoder representations learned from instance-aware supervision to RAW-conditioned demosaicing without introducing semantic prediction at reconstruction time.
    
    \item We introduce DINO-bridged cross-task alignment, which anchors understanding- and reconstruction-conditioned decoder features to a shared frozen representation space and preserves structural organization during task transfer.
    
    \item We validate SegDem with convolutional, Transformer-based, and state-space restoration backbones, demonstrating consistent gains across different architectures and CFA layouts.
\end{itemize}

\section{Related Work}
\label{sec:related_work}

\subsection{Image Demosaicing}
Demosaicing has long been studied as a reconstruction problem under CFA sampling. Classical methods estimate unobserved channel values with interpolation, edge-aware filtering, and cross-channel correlation priors~\citep{malvar2004high}. Learning-based methods replace hand-crafted reconstruction rules with end-to-end networks and often combine demosaicing with denoising to handle realistic RAW degradations~\citep{gharbi2016deep,kokkinos2018deep}. Recent methods further extend this direction to emerging sensor layouts and efficient backbones, including unified Single-, Quad-, and Nona-Bayer demosaicing~\citep{DBLP:conf/iccp/TedlaPZB25}, multispectral demosaicing with dual cameras~\citep{DBLP:conf/iccv/TedlaLYAB25}, Mamba or Transformer designs for different sensors~\citep{DBLP:conf/eccv/ZengLP24,DBLP:conf/nips/PanZCCZX24,DBLP:conf/cvpr/ZhouZLSTCLS25,zhou2025tsanet}, and unified architectures for pixel-bin sensors~\citep{lee2023efficient,kumar2026pixelbin}. Our work differently studies how representations learned through instance-aware structural supervision can be transferred to demosaicing.

\subsection{Large-Scale Generative Models for Low-Level Vision}
Large-scale generative models have recently become strong priors for low-level vision. Stable-diffusion-based methods use pretrained text-to-image models~\citep{DBLP:conf/cvpr/RombachBLEO22,DBLP:journals/corr/abs-2506-15742} for image restoration~\citep{wang2023stablesr,yang2023pasd,lin2023diffbir,wu2024seesr,yu2024supir,DBLP:journals/corr/abs-2508-19154,DBLP:conf/iclr/ZengSCSX25}, while recent one-step and editing-model-based systems improve efficiency or generalization~\citep{wu2024osediff,DBLP:journals/corr/abs-2603-25502}. These works demonstrate the value of high-level priors, but they also shift restoration toward perceptual synthesis. Recent evaluations report over-generation, semantic inconsistency, and uncertain evaluation practices in generative image restoration~\citep{blau2018perception,yin2026howfar,DBLP:journals/corr/abs-2604-03061}. Our work follows the motivation of using high-level priors, via segmentation guidance.

\subsection{Representation Learning}
Representation learning transfers high-level visual knowledge without directly generating image content. DINOv2 provides robust object- and region-level features~\citep{oquab2023dinov2}, while prior restoration and generative-model studies use vision-language priors or representation alignment to improve downstream visual features~\citep{DBLP:conf/icml/RadfordKHRGASAM21,luo2023daclip,yang2025vlmir,DBLP:conf/iclr/YuKJJHSX25,DBLP:journals/corr/abs-2512-10794,DBLP:journals/corr/abs-2605-06388,DBLP:journals/corr/abs-2602-00749}. Related predictive representation methods also learn structure in latent space rather than through direct pixel synthesis~\citep{DBLP:conf/cvpr/AssranDMBVRLB23,DBLP:conf/nips/MoT24,grigore2026jepadepth}. SegDem instead uses frozen DINOv2 tokens as a shared training-time reference for segmentation and demosaicing decoder features, while decoder initialization performs the actual cross-task transfer.

\begin{figure*}[t]
  \centering
  \includegraphics[width=\textwidth]{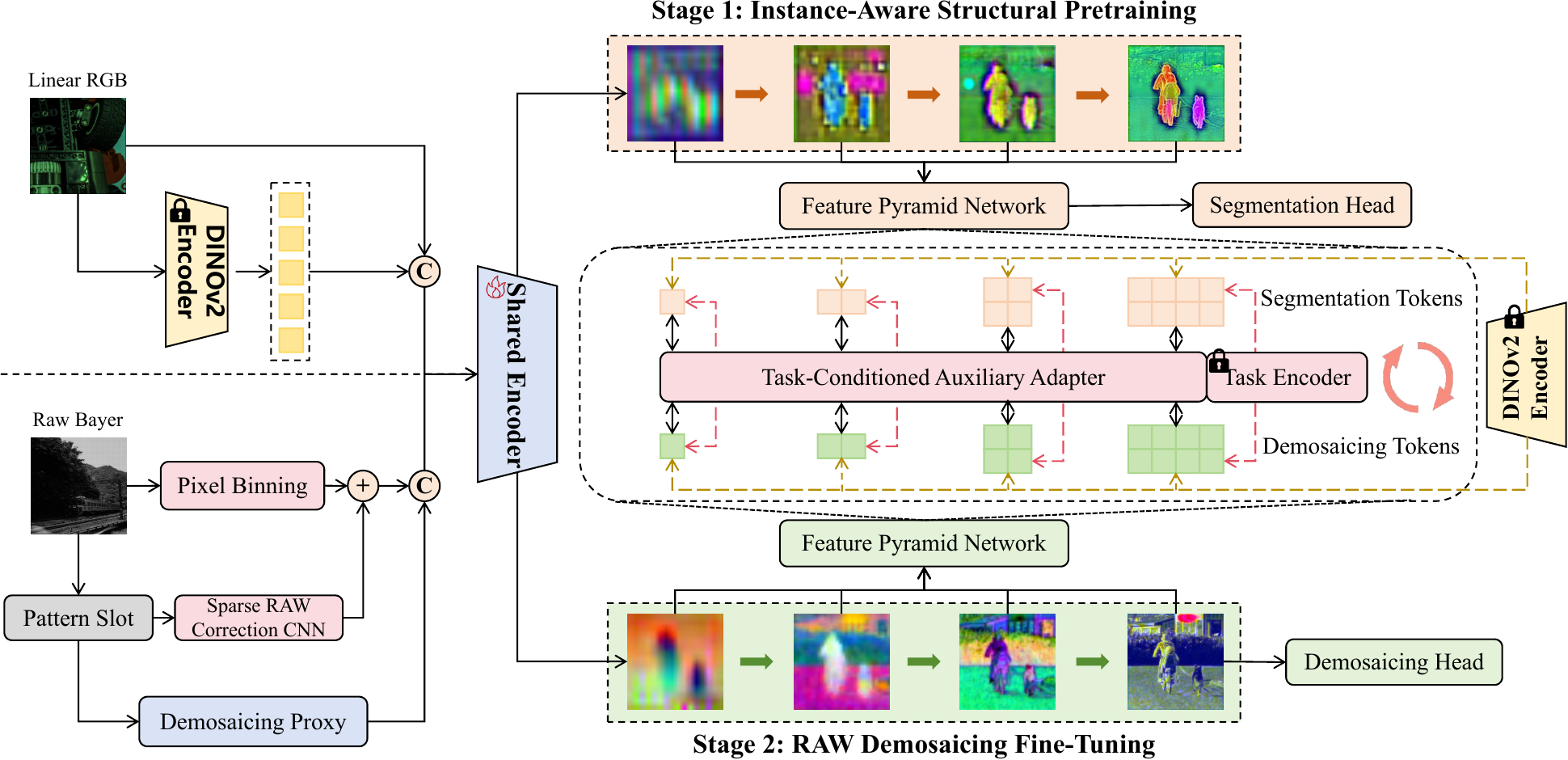}
  \caption{\textbf{Overview of SegDem.} Stage 1 learns region- and boundary-aware representations in a shared RAW-conditioned decoder. Stage 2 transfers the pretrained decoder parameters to demosaicing, while task-conditioned auxiliary features and training-only DINOv2 alignment provide structural supervision without altering the sensor-conditioned output pathway.}
  \label{fig:method}
\end{figure*}

\section{Method}
\label{sec:method}

\subsection{Overview}
\label{sec:method_overview}

SegDem is built on the premise that visual understanding and reconstruction describe the same underlying physical scene and should therefore preserve consistent structural information. We instantiate visual understanding with instance segmentation and transfer its learned structural representation to demosaicing. As illustrated in Figure~\ref{fig:method}, SegDem contains two training stages. Stage~1 pretrains a shared RAW-conditioned decoder with instance-aware structural supervision, while Stage~2 initializes the demosaicing network with the pretrained decoder and fine-tunes it for linear-RGB reconstruction. Decoder initialization performs the cross-task parameter transfer, while task-conditioned auxiliary features and frozen DINOv2 tokens provide a shared structural reference across two stages.

\subsection{Instance-Aware Structural Representation Pretraining}
\label{sec:seg_guided_decoder}

Demosaicing recovers a full linear RGB image $x$ from a sparse CFA observation
\begin{equation}
  y = \mathcal{M}_{p}(x) + n,
  \label{eq:cfa_observation}
\end{equation}
where $\mathcal{M}_{p}$ is the CFA sampling operator and $n$ denotes sensor noise. Because only one color is observed at each sensor location, several cross-channel continuations can be locally consistent with the same RAW measurements. This ambiguity is especially severe around object boundaries, thin structures, and repetitive patterns. We instantiate visual understanding with instance segmentation, whose supervision provides structural information complementary to pixel-level reconstruction.

Stage~1 is designed for representation learning rather than standalone RAW segmentation. The paired clean linear RGB image is used as the image state so that the decoder can focus on region and boundary organization without being dominated by demosaicing errors, while the mosaicked RAW observation is retained as the sensor condition to preserve the interface used in Stage~2. Following Gen2Seg~\citep{DBLP:journals/corr/abs-2505-15263}, instance masks are converted into a dense RGB-coded target that can be predicted by the same image-to-image decoder family used for restoration. This objective encourages pixels belonging to the same instance to share a consistent representation and adjacent instances or background regions to remain separable.

We further use a lightweight instruction adapter to construct task-specific auxiliary views of the shared decoder features. A fixed segmentation prompt and a fixed demosaicing prompt are encoded once by a frozen Qwen2.5-VL~\citep{bai2025qwen25vl} encoder, producing cached embeddings $e_{\mathrm{seg}}$ and $e_{\mathrm{demo}}$. For a decoder feature $h_{\theta}^{\ell}$ at level $\ell$, the task embedding modulates the normalized feature as
\begin{equation}
  \tilde{h}_{\theta,t}^{\ell}
  =
  \left(1+\gamma_{\ell}(e_t)\right)
  \odot
  \operatorname{Norm}\left(h_{\theta}^{\ell}\right)
  +
  \beta_{\ell}(e_t),
  \qquad
  t\in\{\mathrm{seg},\mathrm{demo}\},
  \label{eq:instruction_modulation_main}
\end{equation}
where $\gamma_{\ell}(\cdot)$ and $\beta_{\ell}(\cdot)$ predict channel-wise scale and shift parameters. In Stage~1, $e_{\mathrm{seg}}$ produces instance-oriented auxiliary features for the RGB-coded instance head and representation alignment. The adapter only selects the auxiliary representation objective. Detailed adapter architecture and prompts are provided in Appendix~\ref{app:instruction_adapter}. The Stage~1 objective is
\begin{equation}
  \mathcal{L}_{\mathrm{stage\text{-}1}}
  =
  \lambda_{\mathrm{inst}}\mathcal{L}_{\mathrm{inst}}
  +
  \lambda_{\mathrm{repr}}^{\mathrm{seg}}
  \mathcal{L}_{\mathrm{repr}}^{\mathrm{seg}},
  \label{eq:stage1_loss}
\end{equation}
where $\mathcal{L}_{\mathrm{inst}}$ supervises the RGB-coded instance map and $\mathcal{L}_{\mathrm{repr}}^{\mathrm{seg}}$ anchors the segmentation-conditioned decoder tokens to frozen DINOv2 tokens. After pretraining, the instance head is discarded. Structural information is transferred only through the shared decoder parameters, so Stage~2 does not require segmentation masks, categories, or an additional segmentation output.

\subsection{RAW-Conditioned Demosaicing}
\label{sec:one_step_raw}

Stage~2 receives a proxy linear-RGB reconstruction state and an explicit RAW condition, as shown in Figure~\ref{fig:method}. The proxy is produced by a frozen Jd3Net demosaicing model~\citep{DBLP:conf/iccp/TedlaPZB25}. Since the proxy may already mix color evidence across CFA phases, it is used only as an image-space starting point and is complemented by a separate sensor branch. Pattern-aware CFA packing produces $c_{\mathrm{pack}}\in\mathbb{R}^{4\times H/2\times W/2}$, followed by pixel binning to obtain $c_{\mathrm{bin}}\in\mathbb{R}^{4\times H/4\times W/4}$. A sparse RAW correction module then refines the compact condition before backbone injection, preserving phase-sensitive sensor information. Backbone-specific implementations are detailed in Appendix~\ref{app:backbone_details}.

The reconstruction objective combines linear-domain fidelity, rendered-domain fidelity, perceptual supervision, and edge preservation:
\begin{equation}
\begin{aligned}
  \mathcal{L}_{\mathrm{rec}}
  =&\;
  \lambda_{\mathrm{lin}}
  \mathcal{L}_{\mathrm{char}}(\hat{x},x)
  +
  \lambda_{\mathrm{srgb}}
  \left\|\Gamma(\hat{x})-\Gamma(x)\right\|_1
  +
  \lambda_{\mathrm{1}}
  \mathcal{L}_{\mathrm{lpips}}
  \bigl(\Gamma(\hat{x}),\Gamma(x)\bigr)
  +
  \lambda_{\mathrm{2}}
  \mathcal{L}_{\mathrm{edge}}(\hat{x},x),
\end{aligned}
  \label{eq:rec_loss}
\end{equation}
where $\mathcal{L}_{\mathrm{char}}(a,b)=\frac{1}{|\Omega|}\sum_{u\in\Omega}\sqrt{(a_u-b_u)^2+\epsilon_{\mathrm{char}}^2}$ with $\epsilon_{\mathrm{char}}=10^{-3}$. The fixed PTP operator $\Gamma(\cdot)$ renders camera-linear RGB into sRGB.

\subsection{Structural Representation Anchoring and Objective}
\label{sec:token_regularization}

Decoder initialization transfers the Stage~1 structural representation, but reconstruction fine-tuning may overwrite it with features optimized for local pixel regression. Although segmentation and demosaicing have different objectives, both describe the same underlying scene and should preserve consistent spatial structure. Directly matching their task-specific features is nevertheless undesirable because their feature distributions and optimization objectives differ. We therefore use frozen DINOv2 tokens as a shared structural reference, allowing both task-conditioned representations to be anchored to a common scene representation without directly forcing them to match each other.

\begin{figure*}[t]
  \centering
  \includegraphics[width=\textwidth]{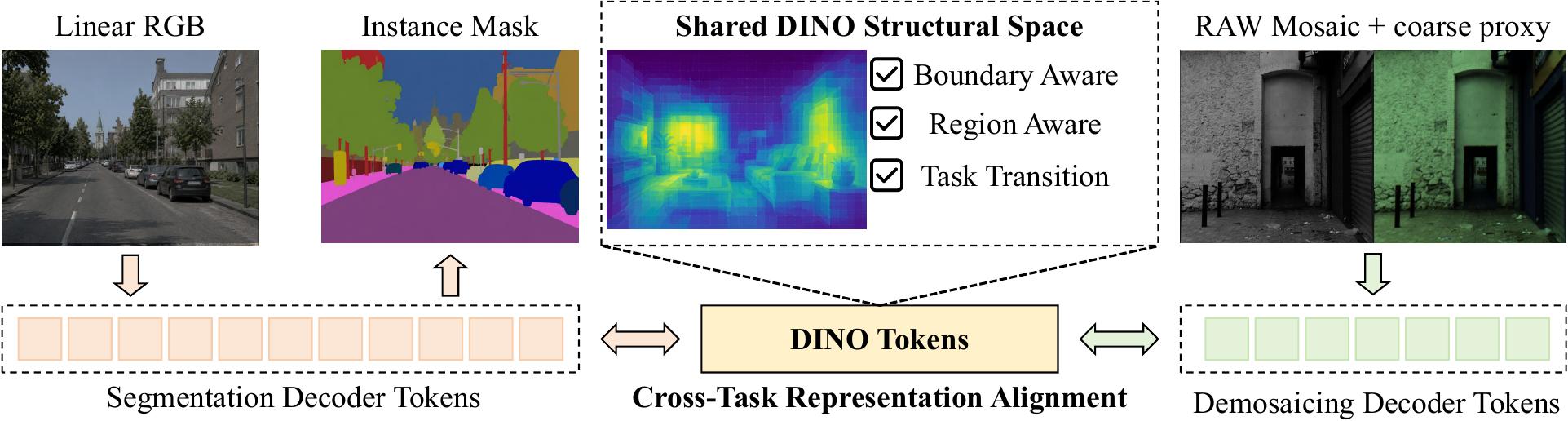}
  \caption{\textbf{Motivation for DINO-bridged structural anchoring.} Segmentation- and demosaicing-conditioned decoder tokens are independently aligned to the same frozen DINOv2 representation space, providing a common structural reference without directly matching task-specific features.}
  \label{fig:transfer}
\end{figure*}

\begin{figure*}[t]
  \centering
  \includegraphics[width=\textwidth]{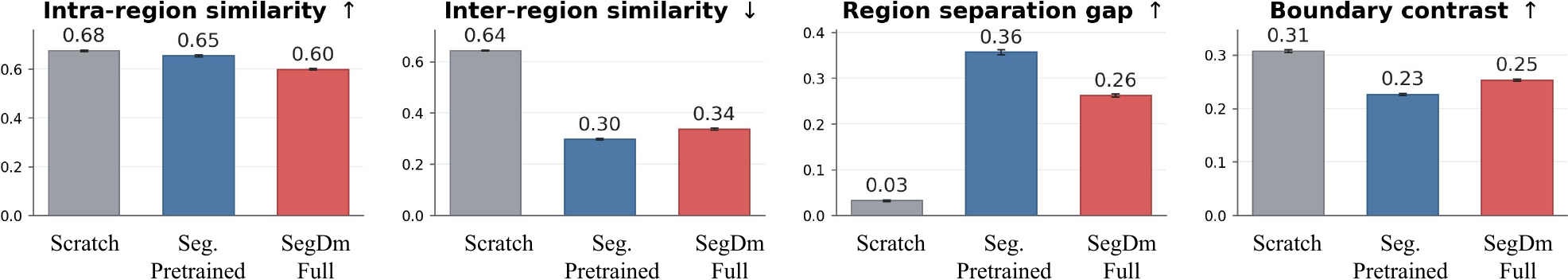}
  \caption{\textbf{Decoder representation analysis.} Scratch training yields local boundary responses but limited separation between regions. Structural pretraining produces more region-discriminative decoder features, and the full model largely preserves this organization after demosaicing fine-tuning.}
  \label{fig:decoder_representation_analysis}
\end{figure*}
Figure~\ref{fig:transfer} illustrates the role of this shared space. In Stage~1, instance supervision introduces region consistency and boundary sensitivity, while $\mathcal{L}_{\mathrm{repr}}^{\mathrm{seg}}$ anchors the resulting segmentation-conditioned tokens to DINOv2. In Stage~2, the transferred decoder is optimized for demosaicing, and $\mathcal{L}_{\mathrm{repr}}^{\mathrm{demo}}$ aligns the demosaicing-conditioned tokens to the same frozen space. Decoder initialization therefore transfers structural knowledge across tasks, while DINOv2 provides a common reference that encourages both representations to preserve consistent scene structure.

For each task $t$, the multi-scale projection head maps the modulated decoder features to dense tokens as $z_{\theta,t}=\mathcal{P}(\{\tilde{h}_{\theta,t}^{\ell}\}_{\ell})$. Frozen DINOv2 target tokens $z_{\mathrm{dino}}$ are extracted from the rendered target $\Gamma(x)$ and reshaped to the same spatial grid. We align tokens at corresponding spatial locations rather than collapsing them into a global feature, because demosaicing errors are strongly location dependent and often concentrate near boundaries or periodic structures. The representation loss is
\begin{equation}
  \mathcal{L}_{\mathrm{repr}}^{t}
  =
  \frac{1}{|\mathcal{M}|}
  \sum_{i\in\mathcal{M}}
  \operatorname{SmoothL1}
  \left(
  \operatorname{LN}(z_{\theta,t}^{i}),
  \operatorname{LN}(z_{\mathrm{dino}}^{i})
  \right),
  \qquad
  t\in\{\mathrm{seg},\mathrm{demo}\},
  \label{eq:repr_loss}
\end{equation}
where $\operatorname{LN}(\cdot)$ denotes token-wise layer normalization and the full token grid contains $N=256$ locations. Layer normalization reduces cross-task scale differences, while Smooth L1 preserves task-specific low-level information.

For each image, we randomly sample a subset $\mathcal{M}$ of tokens for alignment. This selective supervision prevents DINOv2 features from over-constraining the decoder and keeps measurement fidelity governed by the reconstruction objective. The full Stage~2 objective is
\begin{equation}
  \mathcal{L}_{\mathrm{stage\text{-}2}}
  =
  \mathcal{L}_{\mathrm{rec}}
  +
  \lambda_{\mathrm{repr}}^{\mathrm{demo}}
  \mathcal{L}_{\mathrm{repr}}^{\mathrm{demo}}.
  \label{eq:total_loss}
\end{equation}

The representation analysis in Figure~\ref{fig:decoder_representation_analysis} further explains the effect of the proposed transfer and anchoring. Scratch training can produce strong local responses around visible edges, but features from different regions remain highly similar, indicating that edge activation alone does not yield region-discriminative organization. Instance-aware pretraining reduces between-region similarity and enlarges the separation gap. After demosaicing fine-tuning, the full model retains much of this organization, supporting the view that decoder initialization transfers the structural prior and DINO alignment helps preserve it. At inference, the instance head, DINOv2 encoder, representation head, instruction encoder, and random token mask are removed.

\section{Experiments}
\label{sec:experiments}

\subsection{Experimental Setup}
\label{sec:experiment_setup}

We evaluate SegDem on unified Single- and Quad-Bayer demosaicing. We use an image-level 7:1:2 split of LSDIR~\citep{DBLP:conf/cvpr/LiZLCLGZTLDRTG23} for demosaicing training, validation, and testing. COCO~\citep{lin2014microsoft} is used exclusively for instance-aware structural pretraining. The LSDIR test split is disjoint from both training stages and checkpoint selection. RAW inputs are synthesized with the corresponding CFA layouts and Gaussian sensor noise, and the ground-truth targets are linear RGB images. We report reconstruction metrics in linear RGB and rendered sRGB, where sRGB evaluation reflects color and boundary errors after the fixed rendering transform $\Gamma(\cdot)$. The evaluated baselines are Jd3Net~\citep{DBLP:conf/iccp/TedlaPZB25}, BMTNet~\citep{DBLP:conf/cvpr/ZhouZLSTCLS25}, FFTNet~\citep{math14071175}, and ms-demosaic~\citep{DBLP:conf/iccv/TedlaLYAB25}.

This protocol separates the source of structural supervision from the source of RAW reconstruction supervision. COCO provides instance-aware decoder pretraining, whereas LSDIR provides paired RAW/RGB demosaicing supervision. We evaluate the framework with representative convolutional, Transformer, and state-space restoration backbones: EDMUNet~\citep{You_2025_ICCV}, Restormer~\citep{zamir2022restormer}, and MambaIR~\citep{guo2024mambair}. In tables, E, R, and M denote the EDMUNet, Restormer, and MambaIR instantiations of SegDem, respectively. All three instantiations use the same RAW condition, adapter, decoder-transfer, alignment interface, and evaluation protocol.

\begin{table*}[t]
  \centering
  \caption{\textbf{Synthetic Single- and Quad-Bayer demosaicing.} Completed SegDem variants improve over recent baselines; Restormer and MambaIR obtain the strongest fidelity metrics, while EDMUNet gives the best LPIPS. Params(M) reports the full network size.}
  \label{tab:synthetic_demosaic}
  \scriptsize
  \setlength{\tabcolsep}{4pt}
  \resizebox{\textwidth}{!}{
  \begin{tabular}{c|c|ccccc|ccccc}
    \hline
    \multirow{3}{*}{Method} &
    \multirow{3}{*}{Params(M)} &
    \multicolumn{5}{c|}{Single Bayer} &
    \multicolumn{5}{c}{Quad Bayer} \\
    \cline{3-12}
    & &
    \multicolumn{2}{c}{Linear RGB} &
    \multicolumn{3}{c|}{sRGB} &
    \multicolumn{2}{c}{Linear RGB} &
    \multicolumn{3}{c}{sRGB} \\
    \cline{3-4}\cline{5-7}\cline{8-9}\cline{10-12}
    & & PSNR$\uparrow$ & SSIM$\uparrow$ & PSNR$\uparrow$ & SSIM$\uparrow$ & LPIPS$\downarrow$ & PSNR$\uparrow$ & SSIM$\uparrow$ & PSNR$\uparrow$ & SSIM$\uparrow$ & LPIPS$\downarrow$ \\
    \hline
    Jd3Net & 3.20 & 41.1088 & 0.9798 & 30.7496 & 0.8830 & 0.1573 & 39.0379 & 0.9707 & 29.5206 & 0.8656 & 0.2036 \\
    BMTNet & 12.17 & 42.2342 & 0.9823 & 32.7275 & 0.9022 & 0.1334 & 38.6848 & 0.9759 & 30.5340 & 0.8880 & 0.1746 \\
    FFTNet & 43.26 & 44.3537 & 0.9869 & 33.6995 & 0.9178 & 0.1123 & 34.4464 & 0.9211 & 28.8607 & 0.8253 & 0.2267 \\
    ms-demosaic & 10.68 & 44.3198 & 0.9861 & 33.9864 & {0.9197} & 0.1012 & 42.5651 & 0.9769 & 33.0368 & 0.9159 & 0.1211 \\
    \rowcolor{gray!15}\textbf{SegDem(E)} & 78.81 & 44.5753 & 0.9881 & 34.4157 & 0.9176 & \best{0.0832} & 43.1215 & 0.9858 & 33.8543 & 0.9224 & \best{0.0954} \\
    \rowcolor{gray!15}\textbf{SegDem(R)} & 34.28 & \second{45.5654} & \best{0.9908} & \best{35.1883} & \second{0.9312} & \second{0.0891} & \second{43.9275} & \best{0.9884} & \best{34.0871} & \second{0.9249} & \second{0.1061} \\
    \rowcolor{gray!15}\textbf{SegDem(M)} & 5.45 & \best{45.5766} & \second{0.9907} & \second{35.1304} & \best{0.9327} & 0.0951 & \best{43.9385} & \second{0.9883} & \second{34.0355} & \best{0.9261} & 0.1129 \\
    \hline
  \end{tabular}}
\end{table*}

\begin{table*}[t]
  \centering
  \caption{\textbf{SANet-derived evaluation.} The transferred decoder prior generalizes to external pairs, with the largest gains appearing after sRGB rendering and Quad Bayer remaining the harder setting.}
  \label{tab:real_captured}
  \scriptsize
  \setlength{\tabcolsep}{4pt}
  \resizebox{\textwidth}{!}{
  \begin{tabular}{c|ccccc|ccccc}
    \hline
    \multirow{3}{*}{Method} &
    \multicolumn{5}{c|}{RGGB} &
    \multicolumn{5}{c}{Quad Bayer} \\
    \cline{2-11}
    &
    \multicolumn{2}{c}{Linear RGB} &
    \multicolumn{3}{c|}{sRGB} &
    \multicolumn{2}{c}{Linear RGB} &
    \multicolumn{3}{c}{sRGB} \\
    \cline{2-3}\cline{4-6}\cline{7-8}\cline{9-11}
    & PSNR$\uparrow$ & SSIM$\uparrow$ & PSNR$\uparrow$ & SSIM$\uparrow$ & LPIPS$\downarrow$ & PSNR$\uparrow$ & SSIM$\uparrow$ & PSNR$\uparrow$ & SSIM$\uparrow$ & LPIPS$\downarrow$ \\
    \hline
    SANet & 52.7426 & 0.9794 & 42.3979 & {0.9342} & 0.2601 & 52.1435 & 0.9786 & 42.0101 & 0.9211 & 0.2832 \\
    Jd3Net & 53.8407 & 0.9891 & 42.0679 & 0.9434 & 0.2703 & 53.7631 & 0.9876 & 42.1325 & 0.9415 & 0.2631 \\
    ms-demosaic & 55.1885 & 0.9915 & 42.8805 & 0.9473 & {0.2096} & 54.7635 & 0.9902 & 42.6425 & 0.9385 & 0.2863 \\
    BMTNet & 55.3262 & 0.9916 & 43.2509 & 0.9401 & 0.2094 & 52.1638 & 0.9821 & 40.3251 & 0.9311 & 0.3011 \\
    FFTNet & 55.1670 & 0.9917 & 42.9577 & 0.9478 & 0.2516 & 46.3257 & 0.9831 & 36.2578 & 0.9308 & 0.3216\\
    \rowcolor{gray!15}\textbf{SegDem(E)} & 55.4347 & 0.9924 & 43.7238 & 0.9491 & 0.2017 & 55.3304 & \second{0.9926} & 43.7693 & 0.9477 & 0.2586 \\
    \rowcolor{gray!15}\textbf{SegDem(R)} & \second{56.1340} & \best{0.9943} & \second{45.6526} & \best{0.9710} & \best{0.0902} & \best{55.5101} & \second{0.9926} & \second{44.4649} & \second{0.9519} & \second{0.1828} \\
    \rowcolor{gray!15}\textbf{SegDem(M)} & \best{56.1431} & \second{0.9942} & \best{45.6752} & \second{0.9702} & \second{0.0911} & \second{55.4612} & \best{0.9931} & \best{44.5102} & \best{0.9712} & \best{0.1817} \\
    \hline
  \end{tabular}}
\end{table*}

\subsection{Evaluation on Synthetic Benchmarks}
\label{sec:synthetic_results}

Table~\ref{tab:synthetic_demosaic} compares SegDem with recent demosaicing baselines. The completed SegDem variants improve over the strongest baselines on both CFA layouts, showing that the transferred decoder representation is not specific to one sampling pattern. Restormer and MambaIR give the strongest PSNR/SSIM results, while EDMUNet gives the lowest LPIPS. This difference is consistent with different backbone families emphasizing different reconstruction properties: attention and state-space backbones are strong for fidelity-oriented reconstruction, whereas the convolutional decoder benefits more in perceptual boundary and texture quality.

The gain is also more visible after sRGB rendering. Since the fixed rendering transform amplifies chromatic errors and boundary artifacts, improvements in sRGB PSNR/SSIM and LPIPS indicate that the learned structural representation mainly helps in color-ambiguous regions rather than only reducing small linear-domain residuals.

\begin{figure*}[t]
  \centering
  \includegraphics[width=\textwidth]{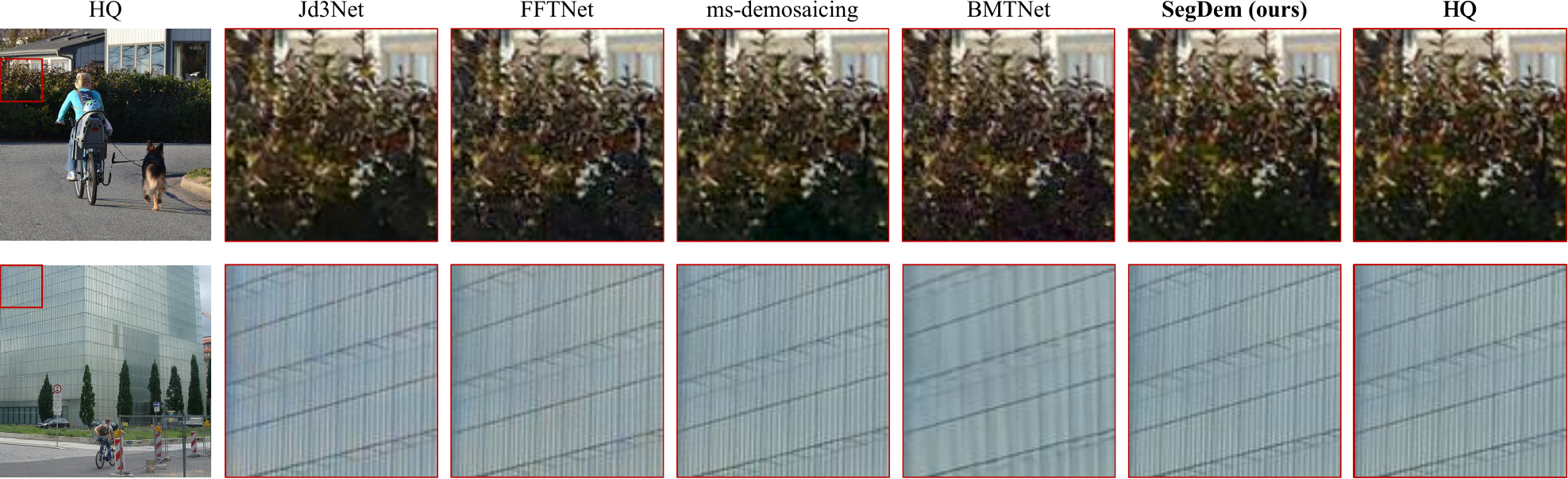}
  \caption{\textbf{Synthetic visual comparison.} Each row uses the same mosaicked observation and compares final demosaicing predictions from representative baselines and SegDem.}
  \label{fig:demosaic_visual_comparison}
\end{figure*}

Figure~\ref{fig:demosaic_visual_comparison} provides a qualitative comparison on representative Single-Bayer and Quad-Bayer RAW inputs. The examples highlight local differences that are not fully captured by global metrics, including color bleeding near object boundaries, zipper artifacts around high-frequency structures, and texture fidelity within coherent regions.

\subsection{Evaluation on SANet-Derived Demosaicing Pairs}
\label{sec:real_results}

We further evaluate on pairs derived from the SANet real image demosaicing dataset~\citep{zhang2022sanet}. The ARQ pixel-shift linear RGB references are cropped to the evaluation size and re-mosaicked with RGGB and Quad Bayer CFA layouts, yielding an external noise-free pure-demosaicing test set that is not used for training or model selection.

Table~\ref{tab:real_captured} shows that the gains also hold in this SANet-derived pure-demosaicing setting. SegDem(R) is consistently strongest, with particularly large improvements in rendered sRGB and LPIPS. This is important because these pairs are not used during training or checkpoint selection; the result suggests that the transferred structural representation improves reconstruction behavior beyond the synthetic LSDIR distribution. Quad Bayer remains more difficult than RGGB for most baselines, but the gap is reduced for SegDem, indicating better robustness to the more ambiguous local CFA grouping.


\subsection{Evaluation on Moir\'e-Hard Datasets}
\label{sec:mit-hard}

Generally, we use the MIT-Hard dataset~\citep{Gharbi:2016:DJD:2980179.2982399} to further examine whether SegDem still maintains good robustness on the Moir\'e-Hard dataset.

\begin{figure*}[t]
  \centering
  \includegraphics[width=\textwidth]{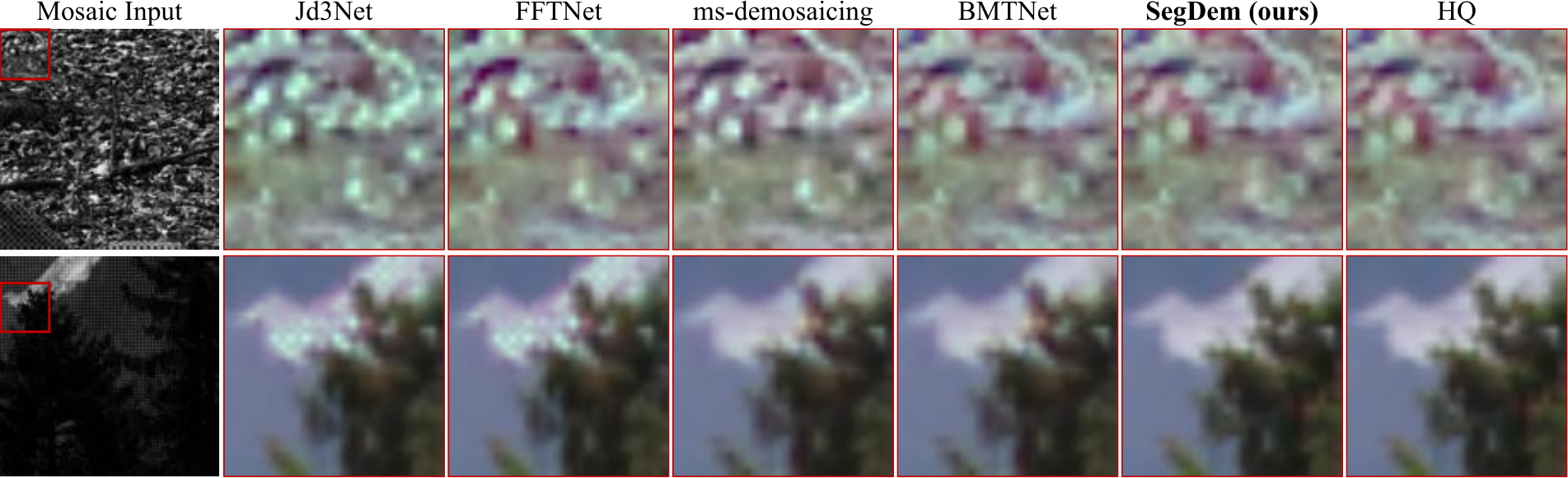}
  \caption{\textbf{MIT-Hard visual comparison.} Each example shows the mosaicked input followed by the final sRGB prediction of each model on the same high-frequency observation.}
  \label{fig:moire}
\end{figure*}

\begin{table*}[t]
  \centering
  \caption{\textbf{MIT-Hard moir\'e-oriented evaluation.} SegDem suppresses residual moir\'e and false-color artifacts on high-frequency content, with the strongest advantage on RGGB and a smaller but consistent margin on Quad Bayer.}
  \label{tab:mit_hard}
  \scriptsize
  \setlength{\tabcolsep}{4pt}
  \resizebox{\textwidth}{!}{
  \begin{tabular}{c|ccccc|ccccc}
    \hline
    \multirow{3}{*}{Method} &
    \multicolumn{5}{c|}{RGGB} &
    \multicolumn{5}{c}{Quad Bayer} \\
    \cline{2-11}
    &
    \multicolumn{2}{c}{Linear RGB} &
    \multicolumn{3}{c|}{sRGB} &
    \multicolumn{2}{c}{Linear RGB} &
    \multicolumn{3}{c}{sRGB} \\
    \cline{2-3}\cline{4-6}\cline{7-8}\cline{9-11}
    & PSNR$\uparrow$ & SSIM$\uparrow$ & PSNR$\uparrow$ & SSIM$\uparrow$ & LPIPS$\downarrow$ & PSNR$\uparrow$ & SSIM$\uparrow$ & PSNR$\uparrow$ & SSIM$\uparrow$ & LPIPS$\downarrow$ \\
    \hline
    Jd3Net & 24.2812 & 0.7792 & 25.3435 & 0.7754 & 0.2443 & 22.4223 & 0.7375 & 24.0604 & 0.7391 & 0.2767 \\
    FFTNet & 24.5212 & 0.7218 & 23.9609 & 0.6934 & 0.3205 & 21.7905 & 0.5246 & 21.2632 & 0.4836 & 0.5158 \\
    ms-demosaic & 27.6128 & 0.8304 & 27.5236 & 0.8081 & 0.2074 & 25.7716 & 0.8040 & 26.1715 & 0.7823 & 0.2427 \\
    BMTNet & 27.3605 & 0.8221 & 27.9221 & 0.8056 & 0.2073 & 25.7552 & 0.8061 & 26.1562 & 0.7811 & 0.2432 \\
    \rowcolor{gray!15}\textbf{SegDem(E)} & 27.8943 & 0.8586 & 27.6161 & \second{0.8257} & \second{0.2040} & 25.9452 & 0.8084 & 26.2920 & 0.7873 & 0.2384 \\
    \rowcolor{gray!15}\textbf{SegDem(R)} & \best{29.4532} & \best{0.8771} & \best{28.9201} & \best{0.8515} & \best{0.1667} & \best{26.8462} & \best{0.8213} & \best{26.6051} & \best{0.8007} & \best{0.2346} \\
    \rowcolor{gray!15}\textbf{SegDem(M)} & \second{28.1354} & \second{0.8673} & \second{28.0143} & 0.8237 & 0.2043 & \second{25.9874} & \second{0.8103} & \second{26.3102} & \second{0.7892} & \second{0.2356} \\
    \hline
  \end{tabular}}
\end{table*}

Table~\ref{tab:mit_hard} shows that the moir\'e-oriented setting is sensitive to both CFA pattern and high-frequency content. Quad Bayer is consistently harder than RGGB because repeated local sampling groups make fine periodic structures more ambiguous. Among standalone baselines, ms-demosaic and BMTNet are the most competitive, but SegDem(R) obtains the best overall reconstruction on both patterns. The advantage is largest on RGGB, while on Quad Bayer, the margin is smaller but remains consistent, which suggests that the structural prior still helps under stronger sampling ambiguity.

\subsection{Ablation Study}
\label{sec:ablation}

\begin{table*}[h]
  \centering
  \caption{\textbf{Effect of structural pretraining across backbones.} Structural pretraining improves the completed backbones, with a larger relative gain for EDMUNet in linear RGB and LPIPS, a stronger sRGB gain for Restormer, and smaller but consistent gains for MambaIR. $\Delta$ denotes Yes--No.}
  \label{tab:pretrain_ablation}
  \scriptsize
  \resizebox{\textwidth}{!}{
  \begin{tabular}{cc|ccccc|ccccc}
    \hline
    \multirow{3}{*}{Backbone} &
    \multirow{3}{*}{Seg. pretrain} &
    \multicolumn{5}{c|}{Single Bayer} &
    \multicolumn{5}{c}{Quad Bayer} \\
    \cline{3-12}
    &
    &
    \multicolumn{2}{c}{Linear RGB} &
    \multicolumn{3}{c|}{sRGB} &
    \multicolumn{2}{c}{Linear RGB} &
    \multicolumn{3}{c}{sRGB} \\
    \cline{3-4}\cline{5-7}\cline{8-9}\cline{10-12}
    & & PSNR$\uparrow$ & SSIM$\uparrow$ & PSNR$\uparrow$ & SSIM$\uparrow$ & LPIPS$\downarrow$ & PSNR$\uparrow$ & SSIM$\uparrow$ & PSNR$\uparrow$ & SSIM$\uparrow$ & LPIPS$\downarrow$ \\
    \hline
    EDMUNet & Yes & 44.5753 & 0.9881 & 34.4157 & 0.9176 & 0.0832 & 43.1215 & 0.9858 & 33.8543 & 0.9224 & 0.0954 \\
    EDMUNet & No & 43.6831 & 0.9848 & 33.5745 & 0.9054 & 0.1174 & 41.6736 & 0.9799 & 32.4886 & 0.9031 & 0.1425 \\
    \rowcolor{gray!10}EDMUNet & $\Delta$ & \textbf{+0.8922} & \textbf{+0.0033} & \textbf{+0.8412} & \textbf{+0.0122} & \textbf{-0.0342} & \textbf{+1.4479} & \textbf{+0.0059} & \textbf{+1.3657} & \textbf{+0.0193} & \textbf{-0.0471} \\
    Restormer & Yes & 45.5654 & 0.9908 & 35.1883 & 0.9312 & 0.0891 & 43.9275 & 0.9884 & 34.0871 & 0.9249 & 0.1061 \\
    Restormer & No & 45.3144 & 0.9891 & 34.1242 & 0.9081 & 0.1007 & 43.7319 & 0.9879 & 33.1773 & 0.9038 & 0.1177  \\
    \rowcolor{gray!10}Restormer & $\Delta$ & \textbf{+0.2510} & \textbf{+0.0017} & \textbf{+1.0641} & \textbf{+0.0231} & \textbf{-0.0116} & \textbf{+0.1956} & \textbf{+0.0005} & \textbf{+0.9098} & \textbf{+0.0211} & \textbf{-0.0116}  \\
    MambaIR & Yes & 45.5766 & 0.9907 & 35.1304 & 0.9327 & 0.0951 & 43.9385 & 0.9883 & 34.0355 & 0.9261 & 0.1129 \\
    MambaIR & No & 45.4913 & 0.9879 & 34.5605 & 0.9301 & 0.0962 & 43.8559 & 0.9883 & 33.5241 & 0.9230 & 0.1144 \\
    \rowcolor{gray!10}MambaIR & $\Delta$ & \textbf{+0.0853} & \textbf{+0.0028} & \textbf{+0.5699} & \textbf{+0.0026} & \textbf{-0.0011} & \textbf{+0.0826} & \textbf{+0.0000} & \textbf{+0.5114} & \textbf{+0.0031} & \textbf{-0.0015} \\
    \hline
  \end{tabular}}
\end{table*}

\begin{figure*}[t]
  \centering
  \includegraphics[width=\textwidth]{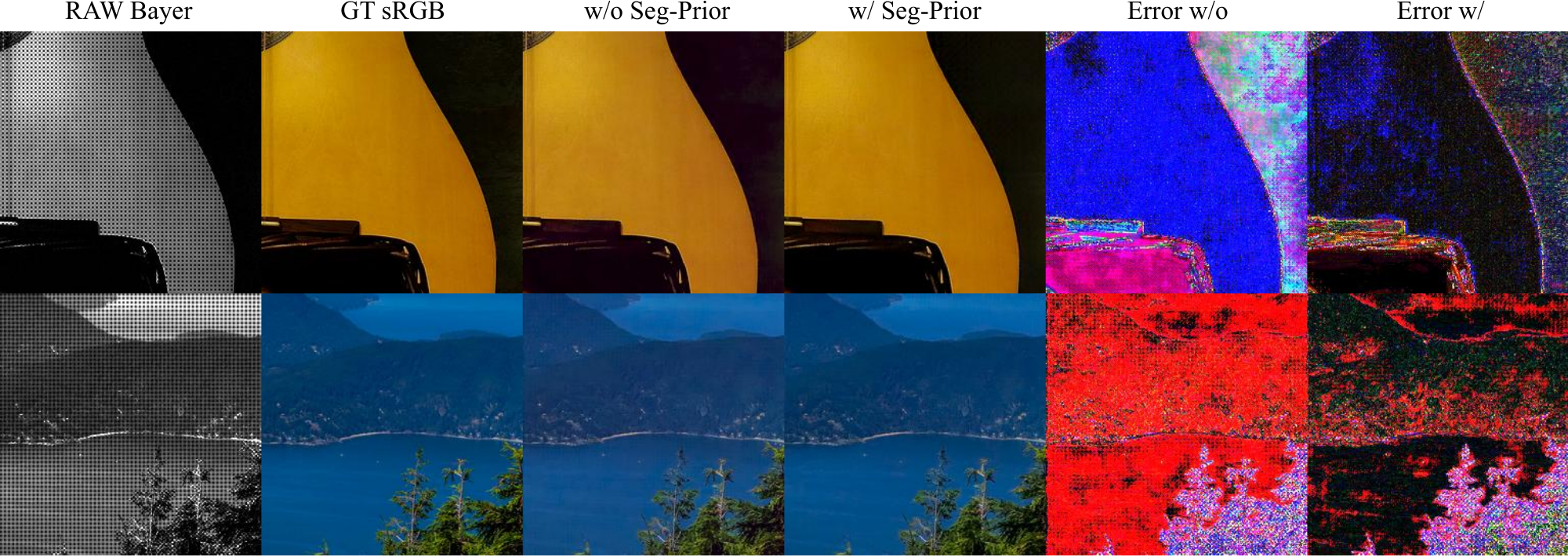}
  \caption{\textbf{Visual ablation of structural pretraining.} The error maps highlight regions where decoder transfer reduces structured color artifacts, especially around boundaries and repetitive textures.}
  \label{fig:seg_prior_visual_ablation}
\end{figure*}

\begin{table*}[t]
  \centering
  \caption{\textbf{Component ablation on the EDMUNet backbone.} The full model is most robust, and the largest drops come from removing the RAW correction branch or DINO-based alignment, confirming that both coarse conditioning and representation anchoring matter.}
  \label{tab:ablation_synthetic}
  \scriptsize
  \resizebox{\textwidth}{!}{
  \begin{tabular}{c|ccccc|ccccc}
    \hline
    \multirow{3}{*}{Method} &
    \multicolumn{5}{c|}{Single Bayer} &
    \multicolumn{5}{c}{Quad Bayer} \\
    \cline{2-11}
    &
    \multicolumn{2}{c}{Linear RGB} &
    \multicolumn{3}{c|}{sRGB} &
    \multicolumn{2}{c}{Linear RGB} &
    \multicolumn{3}{c}{sRGB} \\
    \cline{2-3}\cline{4-6}\cline{7-8}\cline{9-11}
    & PSNR$\uparrow$ & SSIM$\uparrow$ & PSNR$\uparrow$ & SSIM$\uparrow$ & LPIPS$\downarrow$ & PSNR$\uparrow$ & SSIM$\uparrow$ & PSNR$\uparrow$ & SSIM$\uparrow$ & LPIPS$\downarrow$ \\
    \hline
    \rowcolor{gray!15}\textbf{SegDem} & \best{44.5753} & \second{0.9881} & \best{34.4157} & \best{0.9176} & \best{0.0832} & \best{43.1215} & \best{0.9858} & \best{33.8543} & \best{0.9224} & 0.0954 \\
    w/o Demo. Proxy & 40.5308 & 0.9865 & 33.0299 & 0.9088 & 0.0966 & 39.3487 & 0.9732 & 32.7300 & {0.9139} & \second{0.0903} \\
    w/ Bilinear Proxy & 43.5412 & 0.9873 & 34.0301 & \second{0.9130} & 0.0943 & 42.3497 & 0.9813 & 32.9612 & \second{0.9203} & \best{0.0895}\\
    w/o RAW Corr. & 43.6812 & \second{0.9881} & 33.8354 & 0.9035 & 0.1132 & 38.9717 & 0.9742 & 29.4347 & 0.8844 & 0.1829 \\
    w/o DINOv2 Both & 43.1578 & 0.9871 & 32.6838 & 0.8826 & 0.1210 & 42.7350 & 0.9829 & 32.4403 & 0.9029 & 0.1384 \\
    w/o DINOv2 Demo. & 42.2790 & 0.9851 & 33.0503 & 0.9124 & 0.1002 & 41.6277 & 0.9819 & 32.9594 & 0.9118 & 0.1093 \\
    w/o DINOv2 Seg. & 43.6770 & 0.9849 & 33.8177 & 0.9016 & 0.1004 & 41.7929 & 0.9809 & 33.0630 & 0.9043 & 0.1165 \\
    w/o Masked DINO & \second{44.4104} & \best{0.9895} & \second{34.3139} & 0.9044 & \second{0.0939} & \second{43.0261} & \second{0.9844} & \second{33.8345} & 0.9120 & 0.1049 \\
    \hline
  \end{tabular}}
\end{table*}

Table~\ref{tab:pretrain_ablation} isolates the effect of structurally pretrained initialization under the controlled setting described in the caption, and Figure~\ref{fig:seg_prior_visual_ablation} shows the qualitative error maps. The gain is clearer for EDMUNet than for Restormer in linear RGB and LPIPS: on Quad Bayer, EDMUNet improves by +1.4479 dB in linear PSNR and -0.0471 in LPIPS, whereas Restormer improves by +0.1956 dB and -0.0116. This observation is consistent with the hypothesis that the convolutional backbone benefits more from transferred local structural representations when recovering phase-sensitive details. For Restormer, the gain is concentrated in rendered sRGB, suggesting that decoder transfer mainly suppresses errors that become more visible after rendering, such as false colors near boundaries. MambaIR also benefits from structural pretraining, but the margin is smaller. This is expected because the current MambaIR instantiation is closer to a single-scale residual state-space restoration pipeline. The experiments show consistent improvements across different architectures, with larger gains for backbones that expose richer multi-scale decoder representations.

Table~\ref{tab:ablation_synthetic} studies the contribution of individual components on the EDMUNet backbone. The proxy ablations show that a strong reconstruction state is important, but it is not sufficient by itself: replacing Jd3Net with bilinear interpolation keeps reasonable Single-Bayer performance but weakens Quad-Bayer reconstruction, where the CFA ambiguity is larger. Removing the sparse RAW correction module causes the largest Quad-Bayer drop, confirming that the compact RAW condition must preserve phase information rather than only provide a low-resolution color cue. DINOv2-based representation alignment is also important. Removing DINOv2 from both stages reduces reconstruction quality, while removing it only from one stage gives intermediate results. This supports the role of DINO alignment as a training-time anchor for the transferred decoder state. Removing the random token mask keeps competitive linear SSIM, but degrades most PSNR and sRGB perceptual metrics, suggesting that masked token alignment helps regularize decoder features without over-constraining every spatial location.

\section{Conclusion}
\label{sec:conclusion}

This work presents SegDem, exploring the connection between visual understanding and image reconstruction. Our key premise is that both tasks describe the same underlying physical scene and should therefore preserve consistent structural information. We instantiate visual understanding with instance segmentation, using its region- and boundary-aware supervision to pretrain decoder representations that are subsequently transferred to RAW-conditioned demosaicing. A shared frozen DINOv2 representation further anchors structural consistency across the two stages, while explicit RAW conditioning preserves measurement fidelity. Experiments on Single- and Quad-Bayer demosaicing across three different backbones consistently demonstrate the effectiveness and generality of the proposed transfer framework. These results suggest that high-level visual understanding can provide useful structural representations for low-level reconstruction, offering a broader perspective on connecting understanding and reconstruction within a shared visual representation space.

\bibliography{iclr2026_conference}
\bibliographystyle{iclr2026_conference}

\newpage

\newpage

\appendix

\section{Dataset Construction, Training, and Evaluation Details}
\label{app:training_dataset_details}

We use the same paired data format for all demosaicing experiments. Each sample contains a single-channel RAW observation in \texttt{raw\_lq}, a camera-linear RGB target in \texttt{lin\_hq}, and a metadata record that stores the CFA pattern, color correction matrix, channel gains, gamma value, black and white levels, and noise parameters. The metadata is also used by the fixed PTP operator $\Gamma(\cdot)$, so the sRGB losses and sRGB evaluation metrics are computed from the same rendering transform.

\paragraph{Synthetic demosaicing pairs.}
The synthetic demosaicing benchmark is constructed from LSDIR~\citep{DBLP:conf/cvpr/LiZLCLGZTLDRTG23}. We use an image-level 7:1:2 train/validation/test split, so images used for validation or testing are excluded from demosaicing training. We first resize and crop images to $512\times512$, convert display sRGB images to camera-linear RGB with the inverse PTP transform, and then sample the linear RGB image according to the target CFA layout. For Single-Bayer data, we use the standard RGGB arrangement. For Quad-Bayer data, we use a $4\times4$ repeating layout with $2\times2$ same-channel blocks.
The inverse PTP uses a fixed mobile-camera color profile with gamma $2.2$, a fixed $3\times3$ color correction matrix, and fixed channel gains. After mosaicing, we add light signal-independent RAW Gaussian perturbation with $\sigma\sim\mathcal{U}(0,0.01)$. This perturbation is kept small and is shared across all compared methods; the task remains demosaicing from CFA measurements to clean linear RGB. During training, $512\times512$ crops are sampled with even pixel offsets to preserve CFA phase. During validation and testing, center crops are used.

\paragraph{Segmentation pretraining data.}
The segmentation stage uses COCO instance masks under the same image-level 7:1:2 split. Each COCO image and its masks are resized and center-cropped to $512\times512$, converted to camera-linear RGB by the same inverse PTP, and mosaiced into both RGGB and Quad Bayer RAW observations. The instance masks are kept as COCO-style annotations and converted to RGB-coded instance targets for the auxiliary instance-code head. They are used only in segmentation pretraining and are not used as demosaicing targets or during demosaicing inference.

\paragraph{SANet-derived evaluation data.}
For real-reference evaluation, we use the clean split of the SANet real image demosaicing dataset~\citep{zhang2022sanet}. The SANet ARQ files provide pixel-shift camera-linear RGB references. We center-crop each reference image to the evaluation size and re-mosaic it with RGGB and Quad Bayer CFAs to obtain the corresponding RAW observations, producing noise-free paired samples for pure demosaicing evaluation. These SANet-derived pairs are reserved for evaluation only. The metadata records the ARQ source, crop position, CFA pattern, camera white balance derived from the ARQ file, identity color correction matrix, gamma $2.2$, and zero synthetic noise.

\paragraph{Training protocol.}
SegDem is trained in two stages for each reconstruction backbone. In the first stage, the selected RAW-conditioned backbone, RAW correction module, instruction adapters, representation head, and RGB-coded instance head are trained on the COCO RAW instance data using the segmentation instruction. The image state is the clean linear RGB target, while the RAW condition remains packed from the mosaicked observation. The instance loss weight is $\lambda_{\mathrm{inst}}=1.0$ and the segmentation-stage DINO weight is $\lambda_{\mathrm{repr}}^{\mathrm{seg}}=0.1$. The DINO target is extracted from the fixed-PTP rendering $\Gamma(x)$ of the clean linear RGB image. The DINOv2 encoder, the instruction encoder, and the frozen Jd3Net demosaicing proxy are not updated. The optimizer is AdamW with learning rate $5\times10^{-5}$, weight decay $10^{-4}$, gradient clipping at $1.0$, and bfloat16 mixed precision.

In the second stage, the corresponding demosaicing model is initialized from its structurally pretrained checkpoint and fine-tuned on the LSDIR RAW demosaicing pairs for 10 epochs using the demosaicing instruction. The demosaicing-only baseline uses the same demosaicing training data and schedule without segmentation initialization. Training uses AdamW with learning rate $2\times10^{-5}$, zero weight decay, gradient clipping at $1.0$, bfloat16 mixed precision, and distributed data parallel training when multiple GPUs are used. The demosaicing objective uses Charbonnier linear RGB loss, $\ell_1$ sRGB loss after PTP, LPIPS loss in sRGB space, edge loss, and decoder-token DINO alignment with weights $1.0$, $1.0$, $0.2$, $0.1$, and $0.1$, respectively. The noise-conditioning scalar is fixed to $\sigma=0.01$ for the EDMUNet instantiation during demosaicing.

\paragraph{Evaluation protocol.}
Synthetic evaluation uses the held-out test split from the 7:1:2 image-level partition. Validation images are used only for checkpoint selection, and test images are not used in segmentation pretraining, demosaicing training, or model selection. SANet-derived evaluation uses the prepared RGGB and Quad Bayer clean sets only as an external test. We report PSNR and SSIM in linear RGB by directly comparing the predicted linear RGB image with \texttt{lin\_hq}. We also report PSNR, SSIM, and LPIPS in rendered sRGB by applying the metadata-defined PTP operator to both prediction and target. This keeps the linear and sRGB metrics tied to the same paired linear RGB target, while avoiding any learned or image-dependent ISP during evaluation.

\section{Framework Instantiation Details}
\label{app:backbone_details}

SegDem is defined by a common RAW-conditioned reconstruction interface rather than by a single backbone. Each instantiation contains four parts: an image-state branch, a RAW condition branch, task-conditioned auxiliary adapters, and a representation head. Unless otherwise specified, all main experiments use the frozen Jd3Net output as the coarse linear-RGB reconstruction state. SegDem refines this state using the explicit packed-RAW condition and the structurally pretrained shared decoder. The RAW condition branch receives the packed CFA condition described in Appendix~\ref{app:components}. The adapters produce cached task-specific auxiliary features for instance-code supervision and DINOv2 alignment.

Under this interface, the restoration backbones differ only in how they process features and inject the RAW condition. We instantiate SegDem with convolutional encoder--decoder, Transformer-based, and state-space restoration backbones to test whether the transferred decoder representation is tied to a particular architecture.

\paragraph{Convolutional encoder-decoder instantiation.}
The convolutional instantiation uses EDMUNet as the reconstruction backbone. The image branch receives the reconstruction state, while the RAW branch receives the corrected packed CFA condition. The condition is injected into the U-Net feature pathway through convolutional condition projections. Decoder features are exposed through selected decoder blocks for instruction modulation and representation alignment. The network outputs camera-linear RGB in one forward pass. Although this backbone follows a noise-conditioned restoration parameterization, we use a fixed scalar during demosaicing and do not run an iterative sampling process.

\paragraph{Transformer restoration instantiation.}
The Transformer instantiation uses Restormer as the reconstruction backbone. We keep the high-resolution encoder-decoder design and modify only the input, condition, decoder adaptation, and auxiliary representation heads. The image-state branch uses an overlapping patch embedding, followed by four Transformer levels with channel dimensions $48$, $96$, $192$, and $384$. The numbers of Transformer blocks are $[4,6,6,8]$, and the corresponding attention heads are $[1,2,4,8]$. The decoder mirrors the encoder with skip connections, channel reduction after concatenation, three upsampling stages, and four refinement Transformer blocks.

The Transformer condition path receives the four-channel packed RAW condition. A bounded correction module first predicts a residual with three $3\times3$ convolutional layers and a $\tanh$ bound of $0.25$. The corrected condition is resized to the image resolution, embedded by two $3\times3$ convolutions, and added to the patch-embedded reconstruction state with a learnable scalar. This early fusion gives the Transformer blocks access to the sensor constraint while keeping the RAW condition separate from the image-space reconstruction state.

Instruction adapters are inserted after the third, second, and first decoder stages and after the refinement stage. Each adapter follows the FiLM-bottleneck structure in Appendix~\ref{app:instruction_adapter}. The representation head consumes three adapted decoder features, projects them with lateral $1\times1$ convolutions, fuses them in an FPN-style top-down path, and optionally predicts a $16\times16$ grid of 768-dimensional DINO-aligned tokens. During the segmentation stage, the same fused representation also produces an auxiliary RGB-coded instance map. During demosaicing inference, the auxiliary head is not used as an output.

The final Transformer prediction is a residual over the reconstruction state. The output convolution predicts a three-channel correction, which is added to the Jd3Net proxy input and clamped to the valid linear RGB range. This residual formulation makes the backbone focus on demosaicing errors left by the coarse reconstruction state, while the packed RAW condition keeps the reconstruction tied to the measured CFA samples.

\paragraph{State-space restoration instantiation.}
The state-space instantiation uses MambaIR~\citep{guo2024mambair} as the restoration backbone. We keep its shallow convolutional embedding, residual state-space groups, patch embedding/unembedding operators, and final restoration head, and modify only the same interface components used by the other SegDem backbones. The image-state branch receives the coarse linear-RGB reconstruction state. The packed RAW condition is corrected by the bounded RAW correction module, embedded by two $3\times3$ convolutions, and added to the shallow MambaIR feature before the residual state-space groups.

Each residual state-space group exposes a spatial feature map by unembedding the group tokens back to the image grid. We attach the same FiLM-bottleneck instruction adapter to these group features and then re-embed the adapted feature before the next group. The representation head consumes the last three adapted group features, fuses them with the same FPN-style lateral projections, and predicts the $16\times16$ DINO-aligned token grid. When the auxiliary instance-code head is enabled, it is attached to this fused representation in the same way as the other instantiations.

The final MambaIR prediction follows the residual restoration form of the original backbone: the network predicts a correction over the input reconstruction state and outputs camera-linear RGB. Thus the MambaIR variant changes the restoration operator from convolution/attention to selective state-space modeling, while keeping RAW conditioning, task-conditioned auxiliary features, decoder transfer, and inference-time outputs consistent with EDMUNet and Restormer.

\section{Detailed Components of Condition Branches}
\label{app:components}

\begin{figure*}[h]
  \centering
  \includegraphics[width=\textwidth]{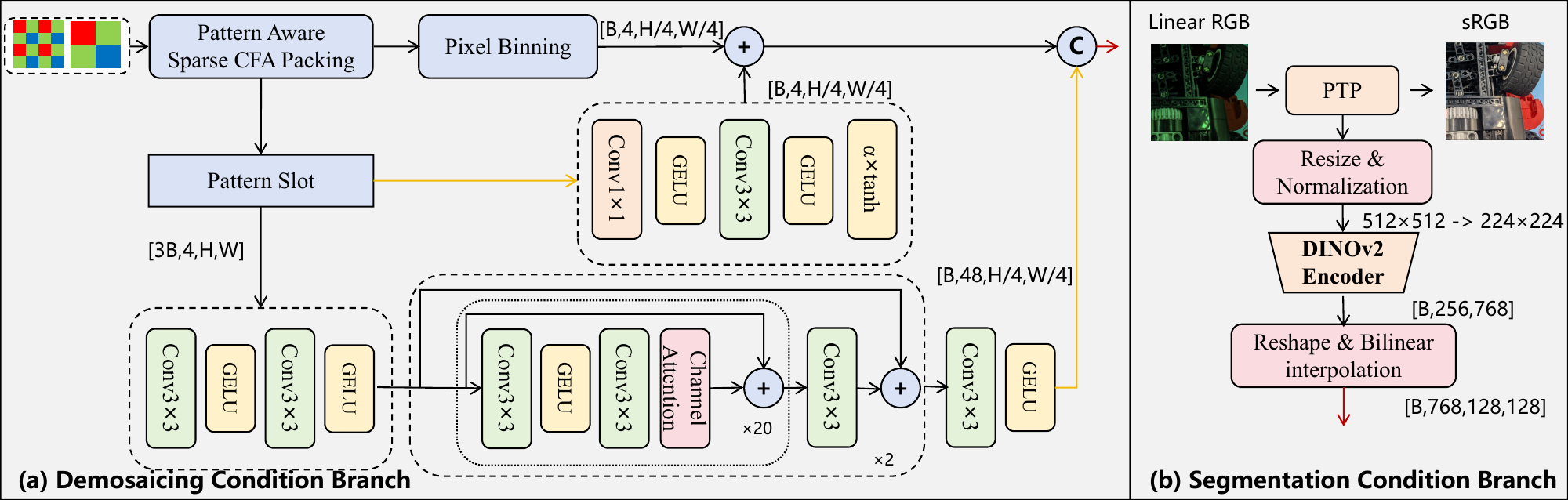}
  \caption{\textbf{Detailed structure of the sensor condition and representation target branches.} The sensor condition branch converts Bayer observations into a packed RAW condition with pattern-aware sparse CFA packing, pixel binning, and sparse RAW correction. The representation target branch renders linear RGB into sRGB with the fixed PTP operator and extracts dense DINOv2 tokens for training-only representation alignment.}
  \label{fig:appendix_condition_branches}
\end{figure*}

Figure~\ref{fig:appendix_condition_branches} shows the detailed condition and representation-target construction used by SegDem. The two branches serve different roles. The sensor condition branch builds a measurement-aware RAW condition for linear RGB reconstruction, while the representation target branch builds training-only semantic targets used to regularize decoder features.

\paragraph{Sensor Condition Branch.}
Given a RAW Bayer observation, the sensor condition branch first applies pattern-aware sparse CFA packing. For Single-Bayer inputs, the RGGB samples are separated into four sparse planes corresponding to $R$, $G_1$, $G_2$, and $B$, giving $c_{\mathrm{pack}}\in\mathbb{R}^{4\times H/2\times W/2}$. For Quad-Bayer inputs, the $2\times2$ same-channel CFA blocks are converted into the same RGGB-like four-plane representation. This design gives different CFA layouts a shared condition format while preserving the location of observed sensor samples.

The packed sparse planes are then reduced by pixel binning to form a compact low-resolution RAW condition $c_{\mathrm{bin}}\in\mathbb{R}^{4\times H/4\times W/4}$. This step is efficient, but it can also blur the difference between CFA phase, local sampling density, and high-frequency sensor evidence. The problem is particularly severe for Quad Bayer inputs. Each color is observed in $2\times2$ same-channel blocks, so binning can collapse several phase-dependent measurements into a similar low-resolution code. The demosaicing backbone then receives a condition that preserves the average color evidence but weakens where and how each sensor sample was observed.

The sparse RAW correction module is designed to recover this missing phase-aware information before the condition is consumed by the backbone. It receives a pattern-slot representation that explicitly encodes CFA phase information, processes it with convolutional blocks and channel attention, and predicts a bounded residual for the binned RAW condition. This residual does not replace the measured RAW values. Instead, it corrects packing and binning artifacts while keeping the condition tied to the observed CFA structure. The corrected RAW condition is concatenated with the projected correction features to produce the final demosaicing condition $c_{\mathrm{raw}}$, which is injected into the one-step RAW-conditioned demosaicing backbone.

The ablation in Table~\ref{tab:ablation_synthetic} confirms the importance of this module. Removing sparse RAW correction has a modest effect on Single Bayer, but it severely degrades Quad Bayer reconstruction, reducing linear PSNR by 4.1498 dB and sRGB PSNR by 4.4196 dB. This gap supports our design choice. Quad Bayer demosaicing is more sensitive to phase loss introduced by compact condition construction, and an explicit phase-aware residual correction is needed to preserve the sensor layout information required for accurate color reconstruction.

\paragraph{Representation Target Branch.}
The representation target branch constructs semantic targets for decoder representation alignment. Since DINOv2 operates on rendered RGB images, the ground-truth linear RGB image is first mapped to sRGB by the fixed PTP operator $\Gamma(\cdot)$. This operator applies metadata-defined channel gains, a $3\times3$ color correction matrix, clamping, and gamma correction. The rendered image is then resized and normalized before being processed by the frozen DINOv2 encoder.

The DINOv2 patch tokens are reshaped into a dense spatial grid and resized to match the decoder-token resolution. These tokens serve as $z_{\mathrm{dino}}$ in Eq.~\ref{eq:repr_loss}. During training, decoder features are projected to $z_{\theta}$ and aligned with $z_{\mathrm{dino}}$, providing a spatial-semantic anchor for decoder features without adding a segmentation output head during demosaicing inference.

\section{Instruction Adapter Details}
\label{app:instruction_adapter}

\begin{figure*}[h]
  \centering
  \includegraphics[width=0.92\textwidth]{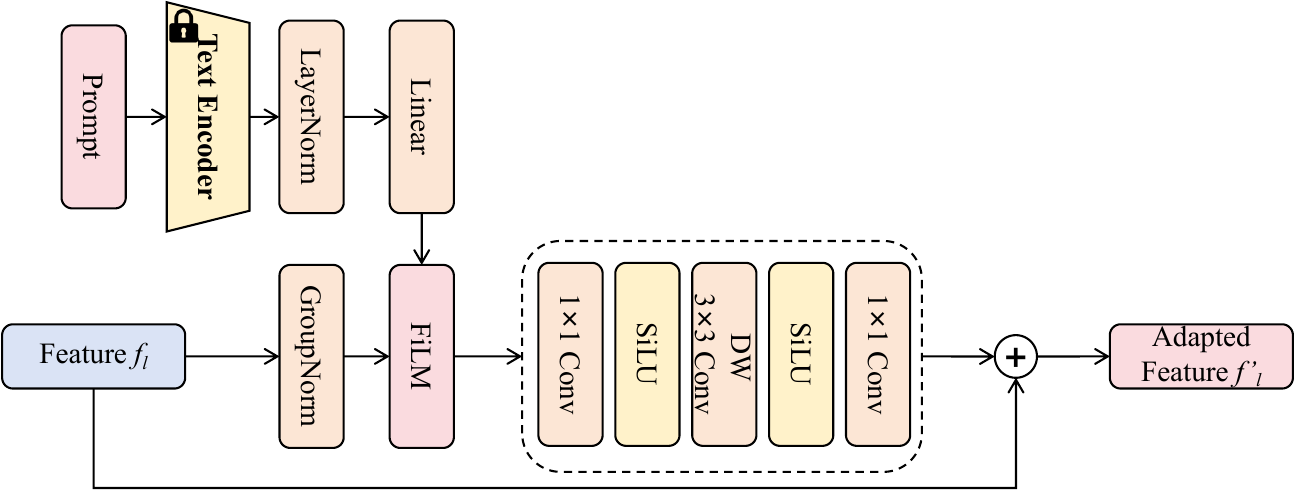}
  \caption{\textbf{Instruction adapter.} A frozen Qwen2.5-VL encoder maps a task prompt into a cached instruction embedding. The embedding is projected to FiLM scale and shift parameters, which produce task-conditioned auxiliary decoder features for instance-code supervision and token alignment.}
  \label{fig:appendix_instruction_adapter}
\end{figure*}

Figure~\ref{fig:appendix_instruction_adapter} illustrates the instruction adapter used to make the shared decoder representation task-aware. The adapter is intentionally lightweight. It does not concatenate language tokens with the RAW input and does not introduce an autoregressive text-to-image generation path. Instead, it converts a fixed task instruction into channel-wise modulation parameters for selected auxiliary decoder features.

We define two task prompts. The demosaicing prompt is ``Recover clean linear RGB from noisy Single-Bayer or Quad-Bayer RAW observations. Preserve edges, color fidelity, and high-frequency details for demosaicing.'' The segmentation prompt is ``Segment every visible COCO object instance from the RAW Bayer scene state. Convert the RAW-conditioned U-Net representation into a mask-centric region representation with instance-level boundaries.'' Each prompt is encoded once by a frozen Qwen2.5-VL encoder~\citep{bai2025qwen25vl}, and the resulting embedding is cached and reused during training and evaluation. Therefore, the vision-language model is not updated by the demosaicing objective and is not repeatedly executed during normal training.

For a decoder feature $f_{\ell}\in\mathbb{R}^{C_{\ell}\times H_{\ell}\times W_{\ell}}$ at level $\ell$, the cached instruction embedding $e_{\mathrm{task}}$ is first normalized and linearly projected to FiLM parameters,
\begin{equation}
  [\gamma_{\ell},\beta_{\ell}] =
  W_{\ell}\mathrm{LN}(e_{\mathrm{task}}),
  \label{eq:instruction_film}
\end{equation}
where $\gamma_{\ell},\beta_{\ell}\in\mathbb{R}^{C_{\ell}}$. The feature modulation is applied after group normalization,
\begin{equation}
  \tilde{f}_{\ell} =
  \mathrm{GN}(f_{\ell})\odot
  \left(1+0.1\tanh(\gamma_{\ell})\right)
  + \beta_{\ell}.
  \label{eq:instruction_modulation}
\end{equation}
The bounded scale term prevents large instruction-induced amplification at the beginning of fine-tuning, while the shift term provides task-specific feature re-centering.

For the auxiliary representation branch, the modulated feature is processed by a residual bottleneck adapter,
\begin{equation}
  f_{\ell}^{\mathrm{out}} =
  f_{\ell} + \alpha_{\ell}
  A_{\ell}(\tilde{f}_{\ell}),
  \label{eq:instruction_adapter}
\end{equation}
where $A_{\ell}$ is a $1\times1$ convolution, SiLU activation, depthwise $3\times3$ convolution, SiLU activation, and another $1\times1$ convolution. The hidden width is $C_{\ell}/4$ with a minimum of 16 channels, and the residual scale $\alpha_{\ell}$ is initialized to $10^{-3}$. We attach adapters to the decoder levels used by the feature pyramid head, namely $16\times16$, $32\times32$, $64\times64$, and $128\times128$ decoder features. These levels cover global layout, object boundaries, and local texture details.

The adapter serves different roles in the two stages. During structural pretraining, the segmentation prompt produces instance-code-oriented auxiliary features consumed by the RGB-coded instance head. This encourages the shared decoder to organize its representation around object regions and boundaries. During demosaicing, the demosaicing prompt produces reconstruction-oriented auxiliary features for DINOv2 token alignment. The sensor-conditioned reconstruction path remains unchanged by the language encoder and continues to be supervised by paired linear RGB targets and constrained by the packed CFA condition. In the implementation, decoder forward hooks cache the adapted features for auxiliary token alignment and task heads. The instruction therefore selects the active auxiliary representation objective without allowing language conditioning to override the measured RAW evidence.

\section{Other Implementation Details}

\subsection{DINOv2 Tokens}
For representation alignment, the target linear RGB image is first rendered by the fixed PTP operator and resized to $224\times224$. We then apply the standard ImageNet normalization used by DINOv2 and feed the image to a frozen DINOv2-base encoder~\citep{oquab2023dinov2}. The class token is discarded, and the remaining patch tokens form a $16\times16$ spatial grid with 768 channels. These tokens are used only as supervision targets. Gradients are not propagated into the DINOv2 encoder.

The reconstruction decoder produces a token grid with the same spatial size and channel dimension through the representation head. For the EDMUNet instantiation, both segmentation and demosaicing stages use masked DINO token alignment. We sample a new random token mask for each batch with mask ratio $0.4$ over the $16\times16$ token grid. The mask is independently sampled per image; selected tokens are included in the Smooth L1 token loss after token-wise layer normalization, while unselected tokens are excluded from the loss rather than zeroed. This masked alignment is used only during training. At inference, the DINOv2 encoder, token targets, and random mask are completely removed.

\subsection{Segmentation Head Details}
\label{app:segmentation_head_details}
The segmentation stage is used to learn decoder representations, not to define an extra output for demosaicing. Inspired by Gen2Seg~\citep{DBLP:journals/corr/abs-2505-15263}, we formulate instance-level supervision as RGB-coded image-to-image prediction rather than conventional instance segmentation. The adapted decoder features are fused by the representation head, and an auxiliary head predicts a three-channel instance-code map $\hat{q}\in[0,1]^{3\times H\times W}$. COCO instance masks are converted to a deterministic RGB target $q$: background pixels are black; each object instance receives a deterministic color code; larger instances are drawn first and smaller instances overwrite overlaps so small objects remain visible.

The instance loss is computed on this RGB-coded map,
\begin{equation}
  \mathcal{L}_{\mathrm{inst}}
  =
  \mathcal{L}_{\mathrm{rgb}}
  +
  \lambda_{\mathrm{rgb\text{-}edge}}\mathcal{L}_{\mathrm{edge}}(\hat{q},q)
  +
  \lambda_{\mathrm{emb}}\mathcal{L}_{\mathrm{emb}}
  +
  \lambda_{\mathrm{dice}}\mathcal{L}_{\mathrm{fg\text{-}dice}}
  +
  \lambda_{\mathrm{focal}}\mathcal{L}_{\mathrm{fg\text{-}focal}}
  +
  \lambda_{\mathrm{bd}}\mathcal{L}_{\mathrm{fg\text{-}boundary}}.
  \label{eq:inst_rgb_loss}
\end{equation}
$\mathcal{L}_{\mathrm{rgb}}$ is a Charbonnier RGB regression loss with separate foreground and background terms, where the background term is weighted by $0.2$. $\mathcal{L}_{\mathrm{emb}}$ is a Gen2Seg-style discriminative embedding loss on the predicted three-channel code: pixels from the same instance are pulled toward their instance mean, different instance means are repelled, and background pixels are encouraged to remain near zero. The edge, foreground Dice, focal, and boundary terms stabilize object extent and boundary learning.
We use $\lambda_{\mathrm{rgb\text{-}edge}}=0.25$, $\lambda_{\mathrm{emb}}=1.0$, $\lambda_{\mathrm{dice}}=1.0$, $\lambda_{\mathrm{focal}}=1.0$, and $\lambda_{\mathrm{bd}}=0.25$.

This head is therefore different from traditional Mask R-CNN-style instance segmentation~\citep{he2017mask}: it has no class prediction, proposal matching, mask queries, or one binary mask per instance. Dice and foreground IoU are computed from the predicted RGB code map by thresholding the maximum RGB channel, $\hat{f}=\mathbf{1}[\max_c \hat{q}_c>0.08]$, and comparing it with the union of COCO instance masks. These metrics are reported only to verify that the segmentation pretraining stage learns meaningful region structure. The segmentation head is discarded during demosaicing.

During segmentation pretraining, the reconstruction state is the paired clean linear RGB image, while the RAW condition remains packed from the mosaicked observation. This prevents the segmentation objective from being dominated by demosaicing errors and focuses the stage on learning object-region and boundary representations. During demosaicing training and inference, the segmentation head is inactive and no mask prediction is produced.

\subsection{Demosaicing Head Details}
The demosaicing head always predicts camera-linear RGB. For EDMUNet, the output is produced by the original convolutional reconstruction head. For Restormer and MambaIR, the output convolution predicts a three-channel residual that is added to the reconstruction state. The final prediction is clamped to $[0,1]$ before losses and metrics are computed. This residual design is used only to stabilize restoration around the proxy reconstruction; the packed RAW condition still provides the sensor constraint.

The reconstruction state is generated consistently for both training and evaluation. We use a frozen Jd3Net demosaicing proxy~\citep{DBLP:conf/iccp/TedlaPZB25}; the sparse RAW input is mapped to the proxy model's expected range. In our experiments this proxy input uses the $[-1,1]$ range, while RAW values and linear RGB targets are stored in $[0,1]$. For Single-Bayer and Quad-Bayer inputs, the appropriate proxy output branch is selected according to the CFA pattern and resized to the RAW resolution if needed.

The sRGB loss and sRGB metrics are computed by the same fixed PTP operator $\Gamma(\cdot)$. We do not train a learned ISP and do not use image-dependent tone mapping during evaluation. Therefore, improvements in sRGB metrics reflect demosaicing errors that remain after a fixed rendering transform, especially false colors and boundary artifacts.

\section{Supplementary Instance Segmentation Evaluation}
\label{app:instance_seg_results}

Although the main paper focuses on RAW demosaicing, we also evaluate the structural pretraining stage on COCO instance masks to verify that the RAW-conditioned backbone learns meaningful region and boundary representations.

\begin{table}[h]
  \centering
  \footnotesize
  \setlength{\tabcolsep}{5pt}
  \caption{\textbf{COCO instance segmentation evaluation.}}
  \label{tab:segmentation_eval}
  \begin{tabular}{lcc}
    \hline
    Strategy & Dice $\uparrow$ & FG-IoU $\uparrow$ \\
    \hline
    Seg. only & 0.7643 & 0.6941 \\
    Demo. init + Seg. & 0.7682 & 0.6969 \\
    \hline
  \end{tabular}
\end{table}

Table~\ref{tab:segmentation_eval} compares two training strategies for the segmentation stage. ``Seg. only'' trains the conditional backbone and the RGB-coded instance head only with the segmentation objective, while ``Demo. init + Seg.'' initializes the same segmentation model from demosaicing pretraining and then fine-tunes it for instance-code supervision. The gain is moderate but consistent, indicating that RAW reconstruction pretraining learns boundary-sensitive and region-aware decoder representations that remain compatible with instance-level semantic grouping.

\begin{figure}[h]
  \centering
  \includegraphics[width=0.85\linewidth]{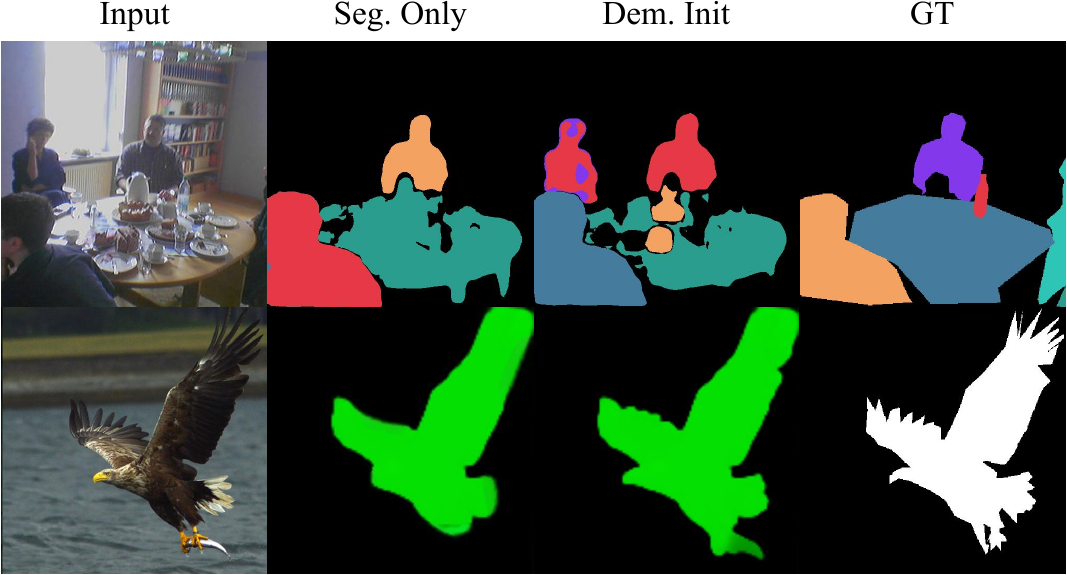}
  \caption{\textbf{Supplementary segmentation visual comparison.} From left to right, we show the rendered input, ground-truth instance masks, segmentation-only prediction, and demosaicing-initialized prediction.}
  \label{fig:segmentation_eval_visual}
\end{figure}

Figure~\ref{fig:segmentation_eval_visual} provides a qualitative view of the same comparison. Both models recover the dominant foreground regions, showing that the RAW-conditioned backbone can support instance-level parsing even when the input comes from mosaicked sensor observations.

\section{Limitations and Future Work}

SegDem provides an initial exploration of the connection between visual understanding and image reconstruction, but several limitations remain. First, we instantiate visual understanding only with instance segmentation, since its region and boundary supervision is particularly relevant to demosaicing. Whether other understanding tasks, such as semantic parsing, depth estimation, or correspondence learning, provide complementary transferable structures remains to be explored. Second, the current framework primarily studies the transfer from understanding to reconstruction. Although our supplementary experiments suggest that reconstruction-pretrained representations can also benefit instance-level understanding, a fully bidirectional framework that jointly exploits ``understanding for reconstruction'' and ``reconstruction for understanding'' is beyond the scope of this work. Finally, our training and evaluation mainly rely on synthetically mosaicked or re-mosaicked RAW observations. Extending SegDem to native RAW data from diverse cameras, sensor noise characteristics, and CFA designs would provide a more comprehensive evaluation of its real-world generalization. These directions may further establish understanding and reconstruction as complementary forms of visual representation learning.

\end{document}

%% file: math_commands.tex
\usepackage{amsmath,amsfonts,bm}

\def\eqref#1{equation~\ref{#1}}

\def\1{\bm{1}}

\DeclareMathAlphabet{\mathsfit}{\encodingdefault}{\sfdefault}{m}{sl}
\SetMathAlphabet{\mathsfit}{bold}{\encodingdefault}{\sfdefault}{bx}{n}

